# Fingerprint Recognition under Missing Image Pixels Scenario

Dejan Brajovic, Kristina Tomovic, Jovan Radonjic
Faculty of Electricity
University of Montenegro
Podgorica, Montenegro
dbrajovic0@gmail.com, kristina.tomovic11@gmail.com, jocoradonjic@gmail.com

*Abstract*— **This work observed the problem of fingerprint image recognition in the case of missing pixels from the original image. The possibility of missing pixels recovery is tested by applying the Compressive Sensing approach. Namely, different percentage of missing pixels is observed and the image reconstruction is done by applying commonly used approach for sparse image reconstruction. The theory is verified by experiments, showing successful image reconstruction and later person identification even if less then 90% of the image pixels is missing.**

## I. INTRODUCTION

In order to determine the identity of a person, the biometric systems are developed with a goal to analyze persons physical and behavioral characteristics. Physical characteristics include fingerprint, facial features, eye pupils, and the behavior ones are signature, voice, stroke, etc.

Fingerprint identification [1]-[7] is the most susceptible biometric method developed in the framework of *dactyloscopy*, discipline of criminal technique, which deals with the study of papillary lines (or papillae), formed in various forms on the cheeks and fingers, palms and feet.

The fingerprint represents an oriented structure made of valleys and reefs [4]. *Minutiae* are the characteristic points of the fingerprint, which direction is typically obtained from local ridge orientation. However, one of the basic assumptions about minutiae is that, based on the knowledge of their layout that changes very often because of its elasticity, it is not possible to get the complete reconstruction of the fingerprint. When scanning an image, it is very important to include the middle section of the printout, as it contains the most information useful for identifying [27].

There is a division of fingerprints in several classes: bow, integral bow, left loop, right loop and circular shape, and are distinguished by the number of singular points and their relative position [27] . The bow classes belong to fingerprints without detected singular points, while the class of the integral bow belongs to the prints in which it is possible to detect a singular point. However, such fingerprints can belong to both the left and right loops, and then they are uniquely determined using their *xy* coordinates. The circular fingerprint includes prints with two singular points. This fingerprint classification allows faster searching, because the scanned image can now be found within a class that is recognized as belonging to it.

The fingerprint image can be corrupted by different type of noise, or during transmission to the recognition software, parts of the image may be corrupted of lost. Here, the possibility to recover lost of corrupted parts is tested (the corrupted part of the image may be considered as a lost information). For that purpose, the Compressive Sensing (CS) is considered [2], [3], [8]-[25]. This is a novel approach to signal sampling and acquisition. It allows signal recovery from significantly less amount of information (i.e. from the under-sampled signal), compared to the traditional approach that is based on Shannon-Nyquist sampling theory [8], [10]. Some a priori defined conditions should be satisfied, such as sparsity of the observed signal and incoherence of the acquisition procedure. The majority of signals that appear in real applications have sparse domain representation. In this domain, there is just a small number of coefficients with large values, while the rest are zero or close to zero. The incoherence can be accomplished by the random acquiring the signal samples form the dense signal domain [8], [10], [12], [16].

The paper is organized as follows. Section II provides basic concepts of the CS together with the algorithm for under-sampled image reconstruction while description of the fingerprint recognition approach is described in Section III. The experimental results and concluding remarks are given in the Sections IV and V, respectively.

## II. THEORETICAL BACKGROUND

*Compressive Sensing*

Compressive Sensing (also known as compressed reading and compressive sampling) is a novel method of signal acquisition and reconstruction. It is based on the principle that, through optimization, the sparsity of a signal can be exploited to sample relatively small number of signal coefficients and keep enough information to recover it [8], [10], [16].

Compressive Sensing (CS in the further text) collects signals that are either sparse or compressible. *Sparsity* is an inherent property of signals whose entire information can be represented only with the help of significant components, compared to the total length of that signal. The signal can have a sparse or compressive representation in the original domain or even in some transformation domains like Wavelet or Fourier transformation.

Formally, the central task of CS is recovering $x \in R^p$ from $y = Cx + n \in R^q$ with $p \ll q$, where *n* denotes additive noise and *p* and *q* the degrees of *R,* the field of real numbers.

If for *x* is assumed that it is a sparse signal then there is:

$$\|x\|_0 := |\{i \in [m] : x_i \neq 0\}| \leq s \quad (1)$$

If there is a case where the vector *x* is not sparse in the standard basis, but there is a $\zeta$ as a sparsifying basis, so that $x = \zeta r$ and *r* is sparse. Now, the next task will be recovering *r* from $f(x) = C\zeta r$, where arbitrarily C is $C \in R^{pxq}$. For noise level $\gamma$, solving the final task in order to get the recovered original vector, there is:

$$\text{minimize } \|r\|_1 \text{ subject to } \|Cr - y\|_2 \leq \gamma \quad (2)$$

where *y* represents the function of vector *x*, as $y = f(x)$

CS allows the faithful reconstruction of the original signal by using some non-linear reconstruction techniques. This method can obtain resolved signals from just a few sensors, with nonadaptive sensing. Because of all these features, CS finds its applications in the areas where number of sensors are limited due to very high cost, or where taking measurements is too expensive. CS today runs an enormous fields such as mathematics and applied mathematics, information theory, computer science, circuit design, signal processing, optical and biomedical engineering, etc.

CS algorithm used for image reconstruction in this work is based on a Total Variation (TV) optimization. The TV is defined on real functions, it is a non-negative function. For signals it refers to an integral absolute gradient. Minimization models are used to reconstruct the image, and it is used to regulate overall variations.

The vector of asquired samples *z* is given as :

$$z = \varphi x = \varphi \omega X = DX, \quad (3)$$

where sparse signal *x* can be represented in the certain transform domain $\omega$, *X* is a transform coefficient and *D* the measurement matrix in CS, while the matrix $\varphi$ denotes random set of the coefficients. Next, solving this problem through minimization over *X* of the regularization function *G(X)* would be :

$$G(X) = \frac{\eta}{2}\|z - DX\|^2 + \alpha T(X), \quad (4)$$

$$T(X) = \|gX\|_{\ell 1} \quad (5)$$

where *T(X)* is TV of the signal *X* and $\alpha \in (0, \infty)$ and *g* is a gradient operator. The optimization is now [30] :

$$\text{min } TV(X) \text{ subject to } z=DX \quad (6)$$

III. FINGERPRINT IDENTIFICATIN METHOD

In the detection function, the scanned image converts to the grayscreen image using rgb2gray function. The same thing works with prints from the base. Edge detection applies on the prints, and using the Prewitt approximation, it is easy to find white and black spots and edges of the prints. This function now performs the detection of white and black spots on the scanned image, then compares them with the white and black spots of the prints from the base. If the matching is greater than 90%, the scanned impression coincided with the corresponding fingerprint.

The edge detection function is to find the borders within the image, in a way of detecting places where the light changes suddenly, which is actually finding points of discontinuity. This technique includes four steps:

1. *Smoothing*
2. *Enhancement*
3. *Detection*
4. *Localization*

In the first step, it is necessary to remove as much as possible noise, taking care that the edges are not damaged or destroyed. In the second step, the imprint is filtered to improve the quality of the edges, and further in the third step, it determines which pixel edges should be labeled as noise and discarded, or be retained. Finally, in the fourth step we determine the exact edge location. Connecting and thinning the edges is the usual procedure in this step.

The quality of edge detection is limited by what an image contains. Sometimes user approximately knows where an edge in the image should be, but it is not shown in the result. So he adjusts the parameters of the program in desire to get the edge detected. However, if the edge he has in mind is not as obvious to the program as some other features are, he will add some 'noise' before the desired edge is finally detected [7]. Edge detecting programs process the image 'as it is'.

Edge detection techniques can be divided into two main groups, Gradient method and Laplacian method, and it supports six detection techniques: Sobel, Prewitt, Roberts, Laplacian of Gaussian, zero-cross and Canni [29].

Edge detection steps using gradient are:
1. Smooth the input image
$$(\hat{f}(x,y) = f(x,y) * G(x,y))$$
2. $\hat{f}_x = \hat{f}(x,y) * M_x(x,y) \rightarrow \frac{\partial f}{\partial x}$

   $\hat{f}_y = \hat{f}(x,y) * M_y(x,y) \rightarrow \frac{\partial f}{\partial y}$
3. $magn(x,y) = |\hat{f}_x| + |\hat{f}_y|$
4. $dir(x,y) = \tan^{-1}(\hat{f}_y / \hat{f}_x)$
5. If $magn(x,y) > T$, then there is possible edge point.

*Edge detection by gradient*

This method finds points with a high gradient in which the gradient intensity is maximal [26].
A gradient is a vector that has intensity and direction, and its mathematical formula is shown below [26] :

$$\nabla f(x,y) = \begin{pmatrix} \frac{\partial f(x,y)}{\partial x} \\ \frac{\partial f(x,y)}{\partial y} \end{pmatrix} \quad (7)$$

The gradient intensity is defined as:

$$|\nabla f(x,y)| = \sqrt{\left(\frac{\partial f(x,y)}{\partial x}\right)^2 + \left(\frac{\partial f(x,y)}{\partial y}\right)^2} \quad (8)$$

Unless it is indicated which method user wants to use, the Sobel method is used by default. In this work the Prewitt method is used.

The Prewitt method finds edges using a Prewitt approximation to a derivative that returns edges at those points where the gradient of the processed image is maximum [27].

*Prewitt edge detections*

The Prewitt operator is a discrete differential operator, developed by Judith M.S. Prewitt. The base of the work of this operator lays in its representation as a corresponding gradient vector or its norm, at each pixel [27]
The Prewitt operator represents the image's convolutions with a small, detachable and integer filter in horizontally and vertically, which is the main reason of the ease and fineness of the calculations. It is also important to know that the approximation of the gradient is relatively raw, especially for the variations of high frequencies in the image.

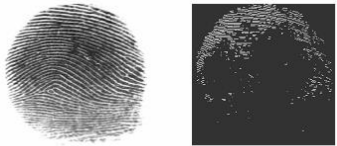

Figure 1. The original print and print after use of these algorithms

Prewitt operator uses 3x3 cores to calculate derivatives approximation. Horizontal and vertical derivatives are calculated as follows [26],[27] :

$$\begin{pmatrix} k_0 & k_1 & k_2 \\ k_7 & [i,j] & k_3 \\ k_6 & k_5 & k_4 \end{pmatrix} \quad (9)$$

If we observe pixels around the pixels (i, j) then the value of a derivative is:

$$H_x = (k_2 + c*k_3 + k_4) - (k_0 + c*k_7 + k_6) \quad (10)$$
$$H_y = (k_6 + c*k_3 + k_4) - (k_0 + c*k_1 + k_2) \quad (11)$$

where the constant c points pixels closer to the center of the mask. If c=1 we get a Prewitt operation:
Horizontal derivate:

$$H_x = \begin{bmatrix} -1 & 0 & 1 \\ -1 & 0 & 1 \\ -1 & 0 & 1 \end{bmatrix} \quad (12)$$

Vertical derivate:

$$H_y = \begin{bmatrix} 1 & 1 & 1 \\ 0 & 0 & 0 \\ -1 & -1 & -1 \end{bmatrix} \quad (13)$$

IV. EXPERMENTAL EVALUATION

For this work the Matlab R2015a is used. Of the functions there are the embedded Matlab functions such as *rgb2gray* and *edge detection* with *Prewitt approximation*..The realization of the program can be described through three main functions - *scann, detection* and *reconstruction*. The first function performs the role of a scanner, reads a fingerprint and invokes the detection function, which checks whether the print is in the database, and shows the one with which the percentage corresponds the most. If the match is over 90%, it is recognized as a successfully detected. The detection function calls the reconstruction function, then reconstructs the imprint and passes it to the detection function, to compare it with examples from the print database.

A. *Example no. 1*

In this work, the fingerprint in the original state is compared with applied CS reconstruction. From the graph (Figure 2.b) it is possible to see that the original printout exists in the base, but after reconstruction with 66% of the available samples, this print is not identified as a known print. With 71% of the available samples, the scanned image can be successfully reconstructed and identified. The example is show in the graph (Figure 2, d). It is important to note that the procedure can work with less available samples as well (67% - 70%), but sometimes fail in recognition. For example, in the case of 69% of the available bounces, out of 10 comparisons we have 4 with the wrong decision, which represents 40% of cases that cannot be ignored.

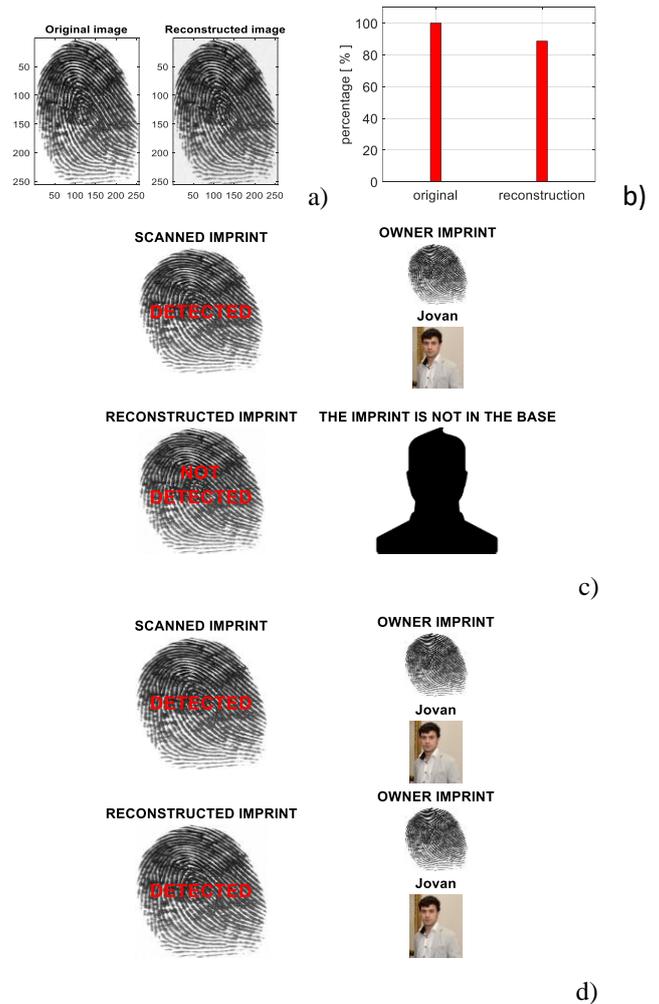

Figure 2. a) original and reconstructed fingerprint; b) identification of the original and reconstructed fingerprint with 66% available samples; c) a percentage matching of the original and reconstructed fingerprint with a fingerprint from the base; d) identification of the original and reconstructed fingerprint with 71% available samples

*B. Example no. 2*

In this example, the print which does not exist in the database is scanned and failed to be identified. After damaging it and the reconstruction with 75% of available samples, the program shows that the print is unknown, which perfectly matches the aim of this experimental work.

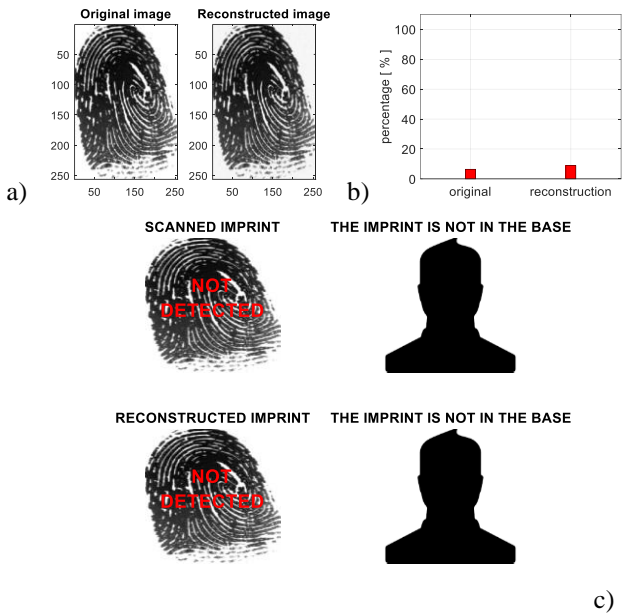

Figure 3. a) original and reconstructed fingerprint; b) a percentage matching of the original and reconstructed fingerprint with a fingerprint from the base; c) identification of the original and reconstructed fingerprint.

V. CONCLUSION

The level of security of biometric devices is directly related to the quality and built-in security mechanisms. Special attention was devoted to testing the biometric system resistance to silicone falsified fingerprints. Namely, testing has shown that the materials needed for the production of silicone fake fingerprints are available everywhere, without the prohibition of purchase or control because it is a widely used material. Instructions for making silicone prints are widely available, mainly over the Internet.

Obviously, this level of security can satisfy most consumers using biometrics in consumer electronics, with intention to remember as few passwords and PINs as possible. However, for professional use, security of data of importance and high risk, for state and security applications, it takes a bit more.